\documentclass{article}
\usepackage{ijcai20}

\usepackage{times}
\usepackage{soul}
\usepackage{url}
\usepackage[hidelinks]{hyperref}
\usepackage[utf8]{inputenc}
\usepackage[small]{caption}
\usepackage{graphicx}
\usepackage{amsmath}
\usepackage{amsthm}
\usepackage{booktabs}
\usepackage{algorithm}
\usepackage{algorithmic}
\usepackage{multirow}
\usepackage[shortlabels]{enumitem}
\usepackage{amssymb}
\title{Characters as Graphs: Recognizing Online Handwritten Chinese Characters \\ via Spatial Graph Convolutional Network}

\author{
Ji Gan$^1$
\and
Weiqiang Wang$^1$\and
Ke Lu$^{2}$\
\affiliations
$^1$ School
of Computer Science and Technology, University of Chinese Academy of Sciences\\
$^2$School of Engineering Science, UCAS
\emails
ganji15@mails.ucas.ac.cn, wqwang@ucas.ac.cn, luk@ucas.ac.cn
}

\begin{document}

\maketitle

\begin{abstract}
  Chinese is one of the most widely used languages in the world, yet online handwritten Chinese character recognition~(OLHCCR) remains challenging. To recognize Chinese characters, one popular choice is to adopt the 2D convolutional neural network~(2D-CNN) on the extracted feature images, and another one is to employ the recurrent neural network~(RNN) or 1D-CNN on the time-series features. Instead of viewing characters as either static images or temporal trajectories, here we propose to represent characters as geometric graphs, retaining both spatial structures and temporal orders. Accordingly, we propose a novel spatial graph convolution network (SGCN) to effectively classify those character graphs for the first time. Specifically, our SGCN incorporates the local neighbourhood information via spatial graph convolutions and further learns the global shape properties with a hierarchical residual structure. Experiments on IAHCC-UCAS2016, ICDAR-2013, and UNIPEN datasets demonstrate that the SGCN can achieve comparable recognition performance with the state-of-the-art methods for character recognition.
\end{abstract}

\begin{figure*}[t]
\centering
\vspace{-0.3cm}
\includegraphics[height=8.5cm]{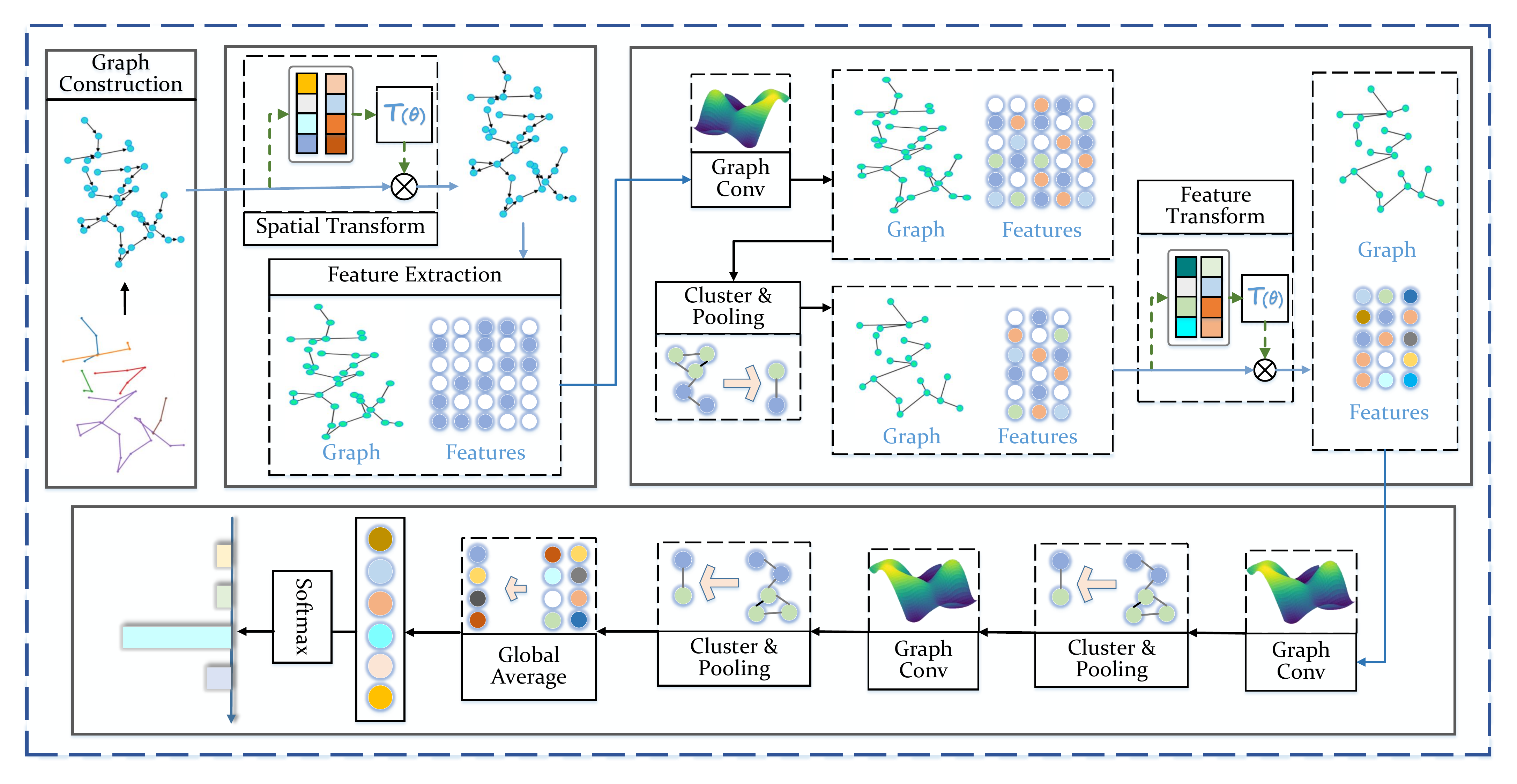}
\vspace{-0.25cm}
	\caption{The overview structure of the proposed graph-based classification for online handwritten Chinese characters. Specifically, (1) a geometric graph is firstly constructed from the normalized handwritten character; (2) after that, deformations of the graph are addressed with spatial transform network~(STN) and then the differentiable feature extraction is conducted; (3) the STN further aligns the shallow-layered features into the latent canonical space; (4) finally, a hierarchical residual structure of graph convolutions is adopted for the final classification.}\label{fig:detailed_gcn}
\vspace{-0.25cm}
\end{figure*}

\section{Introduction}
Chinese characters are among the oldest written languages in the world, which nowadays have been widely used in many Asian countries such as China, Japan, and Korea. Although handwritten Chinese characters can be well recognized by most humans, it remains a challenging issue for computers due to the complex shape structures, a large number of category, and great writing style variations. Therefore, automatic online handwritten Chinese character recognition (OLHCCR) has been widely studied in the past decades for the development of human intelligence.

Traditional methods for OLHCCR generally require to design effective hand-crafted features, upon which classification is performed via traditional machine learning algorithms like the modified quadratic discriminate function (MQDF)~\cite{liu2013online}. With the great success of deep learning techniques, an overwhelming trend is that deep neural networks have gradually dominated the field of OLHCCR~\cite{yin2013icdar,zhang2017online,zhang2018drawing}, outperforming traditional methods with a large margin.

The latest most popular choice for OHLCCR is to adopt the convolutional neural network (CNN) on feature images~\cite{yin2013icdar,zhang2017online}, which converts the problem of character recognition into the image classification. To adopt 2D-CNN for OLHCCR, the most intuitive way is to render handwriting trajectories as static images. To further incorporate the temporal information of handwriting, it is important to utilize the domain-specific knowledge for extracting directional feature images~\cite{zhang2017online}, which improves the recognition accuracy. However, we argue that it is not an ideal solution to represent handwriting as images. Specifically, different from natural images with varied colours and rich content, the character images are typically black-and-white binary-valued, which only provide response values for strokes while leaving large blank areas as background. Therefore, such character images typically contain significant redundant information than natural images. Moreover, transforming 1D trajectories into 2D images increases the data dimension, and thus it may result in huge computational consumption and heavy demand for extra parameters.

Recent advances~\cite{zhang2018drawing,gan2018anew} show that OLHCCR also can be effectively addressed by applying the recurrent neural network (RNN) or 1D-CNN on temporal trajectories, avoiding extracting image-like representations.  However, the RNN remains less amenable to parallelization and suffers low computational speed due to its recurrent computation mechanism; yet the 1D-CNN requires to stack extremely deep layers for learning long-term dependencies of long sequences due to the locality of convolutions. Moreover, since both networks only focus on exploiting temporal information of sequences, it makes them unsuitable for the scenarios that temporal information is lacked or disturbed.

Instead of viewing characters as either images or trajectories, here we propose to represent characters as geometric graphs, naturally retaining both spatial structures and temporal orders. Accordingly, we propose a novel spatial graph convolution network (SGCN) to effectively classify those unstructured graphs as shown in Fig.~\ref{fig:detailed_gcn}. Particularly, we demonstrate that the graph representation has its unique advantages: (1) compared with images, graphs provide a similar natural visual representation, yet graphs keep the essential information with more compact forms. This may lead to much lower computation complexity and fewer model parameters when performing convolutions; (2) compared with trajectories, graphs can store the temporal orders of sequences within the directed edges, but graphs further explicitly reveal the geometric structures. As a result, when the temporal information is lacked, graphs are more discriminative than trajectories and thus can be applied to the more strict scenarios.
\par Our contributions are summarized as follows:
\begin{itemize}[leftmargin=*,itemsep=0.5pt]
\item We propose a compact and efficient geometric graph representation for online handwritten characters, which naturally retains the spatial structures and temporal orders by viewing characters as graphs.
\item We  propose a novel spatial graph convolutional network~(SGCN) for OLHCCR for the first time. The proposed method is largely different from the latest popular methods including the 2D-CNN on feature images, and the LSTM or 1D-CNN on temporal trajectories.
\item The proposed graph-based architecture is fully end-to-end, requiring no human efforts once graphs are constructed. Moreover, it is the first time to verify the effectiveness of SGCN for the large category pattern recognition (nearly four thousands different classes).
\item Experiments are conducted on benchmark datasets (including IAHCC-UCAS2016, ICDAR-2013, and UNIPEN), which demonstrate that the SGCN is very competitive with the state-of-the-art methods for OLHCCR.
\end{itemize}

\section{Related Work}
\paragraph{Online Handwritten Chinese Character Recognition}
OLHCCR has attracted great interests over the past decades, and tremendous achievements have been witnessed in recent years. Specifically, \cite{liu2013online} proposed a standard machine learning-based framework for OLHCCR, which first extracts hand-crafted features from the normalized characters and then performs classification with the modified quadratic discriminate function (MQDF). As an alternative to MQDF, \cite{qu2018air} recently introduced the sparse representation based classification for air-written Chinese characters.

\par With the great impact of deep learning, the CNN has been gradually applied for OLHCCR~\cite{yin2013icdar}, which regards character recognition as image classification. Furthermore, \cite{zhang2017online} proposed to incorporate traditional domain-specific knowledge for extracting feature images, thus further improving the recognition performance of CNN. However, this method faces two problems: (1) the complex domain-specific knowledge for extracting feature images and (2) demand for huge computation and large parameters after increasing the data dimension.
\par Instead of transforming online trajectories into image-like representations, \cite{zhang2018drawing} proposed to directly apply the recurrent neural network (RNN) on temporal trajectories, avoiding the complex domain-specific knowledge for  extracting feature images. Unfortunately, the RNN remains less amenable to parallelization and suffers low computational speed for long sequences. Instead, \cite{gan2018anew} proposed to directly classify temporal trajectories with the 1D-CNN, which empirically runs faster than the RNN for long sequences. However, to exploit long-term dependencies of sequences, the 1D-CNN requires to stack extremely deep layers due to the locality of convolutions. Moreover, both the RNN and 1D-CNN only focus on sequential learning, and thus they are unsuitable for the scenarios that temporal information is lacked (such as the offline HCCR).

\par Rather than viewing characters as static images or temporal trajectories, here we propose to represent characters as geometric graphs. Accordingly, a spatial graph convolutional network (SGCN) is proposed for character recognition for the first time, addressing OLHCCR in a novel perspective.

\paragraph{Convolution on Graphs} Although the traditional CNN has achieved great success in many domains, it is limited to processing images with regular grid structures. Instead, recent advances have generalized convolutions on irregular unstructured graphs, which can be divided into two broad categories: the spectral methods~\cite{kipf2016semi,Micha2016fastgcn,yu2018spatio} and spatial methods~\cite{monti2017geo,fey2018splinecnn,wang2019dynamic}. The spectral methods typically apply the Fourier transform on graphs, converting the graph convolution into multiplication in the spectral domain. However, the spectral methods require the input graphs to have the identical structures; and worse still, the spectral convolution only encodes the connectivity of nodes but ignores their geometric information. On the contrary, the spatial methods perform convolutions to aggregate local  neighbourhoods of graph nodes in the spatial domain via the weighted sum, where the edge weights are dynamically calculated depending on the local geometry. Therefore, our work follows the spirit of spatial graph convolutions to fully exploit the spatial geometric properties of character graphs. Moreover, we are among the first to verify the effectiveness of spatial graph convolution for the large category pattern classification (which involves nearly four thousand classes).

\section{Methodology}

Our goal is to represent online handwritten characters as sparse geometric graphs and then perform convolutions on the constructed graphs for the final classification. The overview of the spatial graph convolutional network~(SGCN) for character recognition is illustrated in Fig.~\ref{fig:detailed_gcn}, and the details of each part will be described in the following section.

\subsection{Graph Construction}
\begin{figure}[t]
\centering
\includegraphics[height=2.2cm]{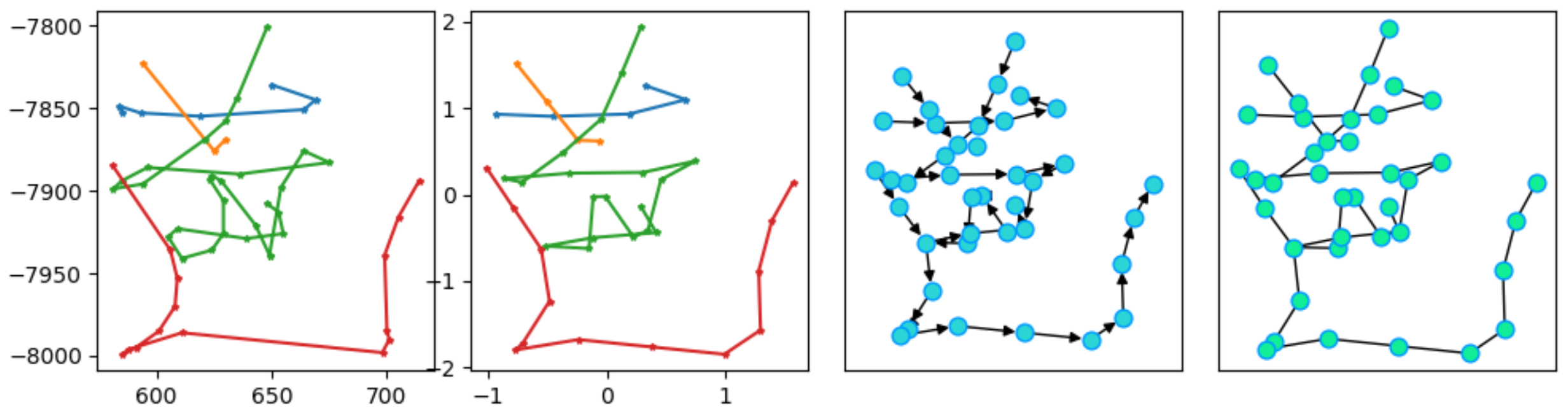}
\vspace{-0.2cm}
	\caption{Graph construction. From left to right: original \& normalized trajectories, and directed \& undirected graphes respectively.}\label{fig:graph_construction}
\end{figure}

Handwriting typically contains rich diversity of writing styles, and it also suffers from large variations in spatial sizes as well as locations. Therefore, it is important to normalize handwritten characters to reduce those variations for extracting reliable features. Similar to ~\cite{zhang2018drawing}, each character is firstly normalized into a standard $xy$-coordinate system with its shape unchanged. Moreover, each handwriting stroke is further re-sampled into the same interval.

After the normalization, we obtain a point sequence of the length $N$ with absolute coordinates as $[[x_1, y_1],\dots,[x_i, y_i],\dots,[x_N, y_N]]$. As shown in Fig.~\ref{fig:graph_construction}, we propose to construct a direct geometric graph $\mathbb{G}=(\mathbf{V}, \mathbf{E}, \mathbf{P})$ from the given point sequence of length $N$, where $\mathbf{V}$ denotes the node-set, $\mathbf{E}$ denotes the edge-set, and $\mathbf{P}$ denotes the coordinate-set of all nodes. In the graph, each point is corresponding to a node entity $v_i$, and the node-set $\mathbf{V}=\{v_i|i=1,\dots,N\}$ includes all the sampling points in the given sequence, where  $\mathbf{V} \in \mathbb{R}^{N\times 1}$. Accordingly, the absolute coordinate $[x_i, y_i]$ of each node $v_i$ is stored in $\mathbf{P}=\{p_i|i=1,\dots,T\}$, where $\mathbf{P} \in \mathbb{R}^{N \times2}$. Moreover, the edge-set $\mathbf{E}=\{(i, j)|v_j \rightarrow v_i\}$ is constructed depending on the node connectivity, where $v_j \rightarrow v_i$ denotes that $v_j$ points to $v_i$ and $\mathbf{E} \in \mathbb{|R|}^{|\mathbf{E}|\times2}$. Additionally, $|\mathbf{E}| \ll N^2$ in a sparse graph. Lastly, the remaining issue is how to define effective features for each node $v_i$, and we will discuss this next.

\subsection{Feature Extraction Module}
Once the directed graph $\mathbb{G}=(\mathbf{V}, \mathbf{E}, \mathbf{P})$ is constructed from the trajectory, the spatial structure of character is stored in the position-set $\mathbf{P}$ and the temporal information is embedded in the set of directed edges $\mathbf{E}$. Intuitively, we define features of node $v_i$ to retain both spatial  and temporal information, i.e.,
\begin{flalign*}
f({v_i})=[x_i, y_i, \Delta x_i, \Delta y_i, \sin \theta_i, \cos \theta_i],
\end{flalign*}
where $[x_i, y_i]$ denotes the absolute coordinate, $[\Delta x_i,\Delta y_i]$ denotes the writing offset, and $\theta_i$ denotes the writing direction of node $v_i$. Particularly, we regard $[x_i, y_i]$ as the spatial features and $[\Delta x_i, \Delta y_i, \sin \theta_i, \cos \theta_i]$ as the temporal features.
\par It should be noted that all those features can be computed from the directed graph $\mathbb{G}$ with differentiable operations, and thus we can integrate the feature extraction into the back-propagation procedure rather than the pre-processing stage. After that, we add self-connections for all the nodes and further transform the directed graph into the undirected one. This eventually leads the graph convolution to fully incorporate information from the node $v_i$ and all its neighbourhoods.

\begin{figure}[t]
\centering
\includegraphics[height=2.cm]{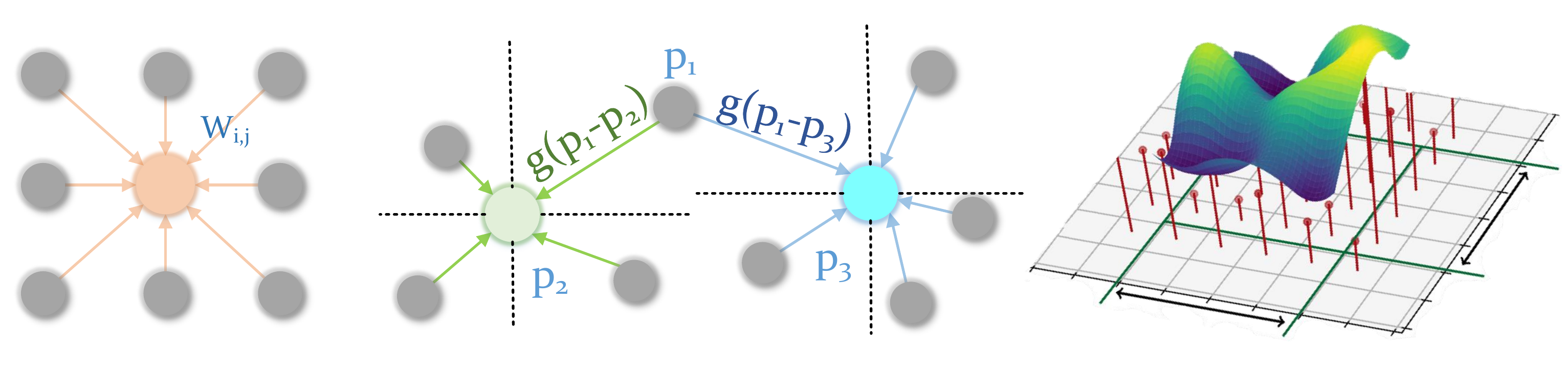}
\vspace{-0.2cm}
	\caption{Convolution operations: (left) Conventional convolution, (middle) Dynamic graph convolution, (right) Spline convolution kernel. In addition, the self-connection is ignored for simplicity.}\label{fig:graph_convs}
\end{figure}

\subsection{Spatial Graph Convolution}~\label{subsec:sgn}
Formally, the convolution at the position $p=[x, y]$ of a 2D grid image (or a single feature map) $\mathbf{F}$ can be defined as
\begin{equation}\label{eq:1}
  \mathbf{F}*g(x, y)=\iint \limits_{(\delta_x, \delta_y) \in \mathbf{G}}{\mathbf{F}(x - \delta_x, y-\delta_y)\mathbf{g}(\delta_x, \delta_y)} {d \delta_x d\delta_y},
\end{equation}
where $\mathbf{G}$ denotes the local region that centred around the position $p$ of $\mathbf{F}$, $[\delta_x, \delta_y]$ denotes the coordinate offset to the center position $p$, and $\mathbf{g}$ denotes the convolutional kernel. As shown in Fig~\ref{fig:graph_convs}~(left), when $\mathbf{F}$ is represented as the irregular 2D grid structure (i.g. an image), the convolution kernel $\mathbf{g}$ is easy to be implemented with a $k \times k$ learnable matrix, where $k$ is well-known as the kernel size.

\par However, it is not straightforward to directly apply the convolution to the unstructured graphs, since the structures of local regions at different positions in a graph can be largely different (i.g. Fig~\ref{fig:graph_convs}~(middle)). Considering that the convolution aggregates information from the local neighbourhoods, the convolution at the node $v_i$ of graph $\mathbb{G}$ can be defined as
\begin{equation}\label{eq:2}
  \mathbf{F}*g(v_i)=\frac{1}{|\mathcal{N}(v_i)|} \sum_{v_j \in \mathcal{N}(v_i)}f(v_j)\mathbf{g}(u(i,j)),
\end{equation}
\vspace{-0.05cm}
where $f(v_j)$ denotes features of the node $v_j$, $\mathbf{F}=\{f(v_i)|i=1,\dots,N\}$, $\mathcal{N}(v_i)$ denotes the neighbourhood nodes of $v_i$, and $u(i,j)=(p_j - p_i)$ denotes the coordinate offsets from $v_j$ to $v_i$. Particularly, the term $\frac{1}{|\mathcal{N}(v_i)|}$ normalizes the cardinality of the corresponding subset. In graphs, the coordinate offsets $u(i,j)$ can be any possible values at different local regions; therefore, the remaining issue is how to define the spatial convolution kernel $\mathbf{g}$ for handling the irregular local structures.

\par Institutively, the basic idea is to fit a continuous function (i.e. a curve surface) depending on the distribution of $u(i,j)$ to approximate the convolution kernel $\mathbf{g}(\mathbf{u})$, where $\mathbf{u}=\{u(i,j)|i=1,\dots,N; v_j \in \mathcal{N}(v_i);u(i,j)=(p_j - p_i)\}$. Fortunately, many spatial convolution kernels recently have been proposed~\cite{monti2017geo,fey2018splinecnn,wang2019dynamic}, and among them, \cite{fey2018splinecnn} is demonstrated to perform the best in many scenarios. Hence, the spline graph convolution kernel (as shown in Fig~\ref{fig:graph_convs}~(right)) is adopted in our work, which can be formally defined as
\begin{equation}\label{eq:2}
  \mathbf{g(u)}=\sum_{p \in \mathbf{P}}{w_p \mathbf{B}_p(\mathbf{u})},
\end{equation}
where $\mathbf{P}$ is the Cartesian product of the ${B}$-spline bases, $\mathbf{B}_p$ is the ${B}$-spline function, and $w_p$ is the trainable parameters to control the height of ${B}$-spline surface. More details of ${B}$-spline convolutional kernel can refer to~\cite{fey2018splinecnn}.

\subsection{Spatial Transform Network}
Since the geometric structure $\mathbf{P}$ of graph $\mathbb{G}$ is corresponding to the absolute coordinates of all the points, the node features as well as graph convolutions are not invariant to the certain geometric transformations of graphs (i.g. the rotation, scaling, and translation). Inspired by~\cite{qi2017pointnet}, the spatial transform network~(STN) is utilized to align the input graph into a canonical coordinate system before feature extraction. \par Specifically, we first encode the geometric structure $\mathbf{P} \in \mathbb{R}^{|N|\times2}$ into a high-dimensional feature vector $\mathbf{h} \in \mathbb{R}^{1 \times d}$as
\begin{equation}\label{eq:3}
  \mathbf{h}=\max \mathcal{M}(\mathbf{P}),
\end{equation}
where $\mathcal{M}$ denotes a multi-layered perceptron~(MLP). After that, we estimate a transform matrix $\mathbf{T}$ based on the embedded vector $\mathbf{h}$. In our task, we further constrain $\mathbf{T}$ to be the similarity transform for simplicity and stability, i.e.,
\begin{equation}\label{eq:4}
\mathbf{T}=\begin{bmatrix} s \cdot \cos \theta & - s \cdot \sin \theta &  \Delta x \\s \cdot \sin \theta & s\cdot \cos \theta &\Delta y\end{bmatrix},
\end{equation}
where $\theta$ denotes the rotation angle, $s$ denotes the scale, and $[\Delta x, \Delta y]$ denotes the transition of $xy$-coordinates, which can be calculated as
\begin{eqnarray}\label{eq:5}
  [\tilde{\theta}, \tilde{s}, \Delta x, \Delta y]&=& \mathcal{M}(\mathbf{h}), \\
  \theta = \pi \cdot \tanh \tilde{\theta}&,&   s = \exp(\tanh(\tilde{s})),
\end{eqnarray}
Therefore, the aligned geometric structure $\mathbf{\tilde{P}}$ is computed as
\begin{equation}\label{eq:6}
\mathbf{\tilde{P}}=\begin{bmatrix} \mathbf{P}& \mathbf{1}\end{bmatrix}  \mathbf{T}',
\end{equation}

Furthermore, This idea can be further extended to the node features $\mathbf{F} \in \mathbb{R}^{|N|\times d}$ as
\begin{eqnarray}\label{eq:6}
\mathbf{h}&=&\max \mathcal{M}(\mathbf{F}),\\
\mathbf{T}&=& \mathcal{R}( \mathcal{M}(\mathbf{h})),\\
\mathbf{\tilde{F}}&=&\mathbf{F}\mathbf{T},
\end{eqnarray}
where $\mathcal{R}$ reshapes the 1-dimensional $d\times d$ vector into a square matrix, and the feature transform matrix $\mathbf{T}$ is not constrained here. Ideally, this STN aligns the intermediate features into a latent canonical space and thus helps learn the geometric invariant features, essentially benefiting the final classification.

\subsection{Hierarchical Residual Structure}
\begin{figure}[t]
\centering
\includegraphics[height=5.2cm]{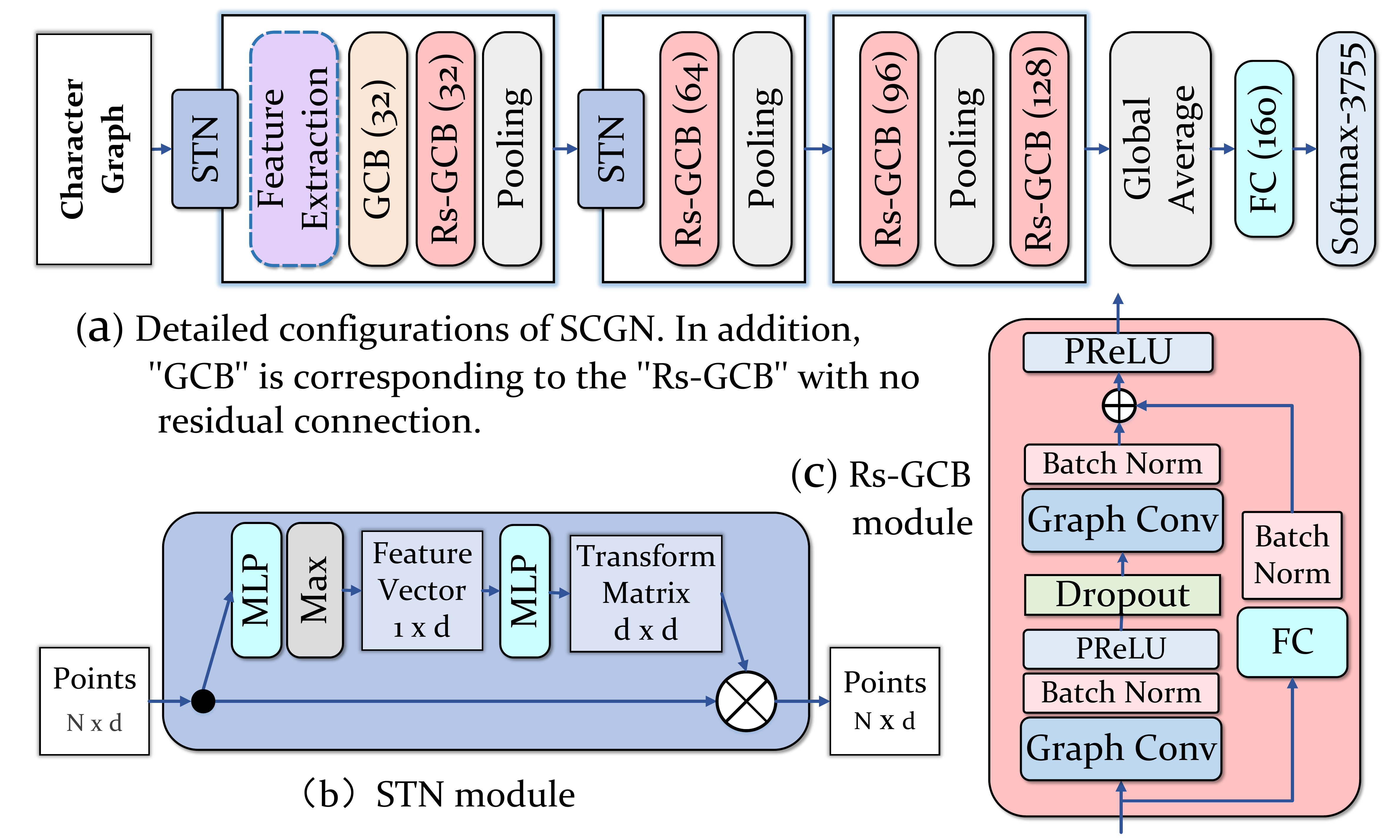}

\vspace{-0.1cm}
	\caption{Configurations of our SGCN for OLHCCR. In the figure, ``Rs-GCB~(32)'' denotes a Rs-CGB module with 32 output channels, ``FC~(160)'' denotes a fully-connected layer with 160 output neurons, and ``MLP'' denotes a three-layered perceptron respectively.}\label{fig:gcn_arch}
\vspace{-0.2cm}
\end{figure}

Traditional graph neural networks~(GNNs) are inherently flattened and cannot learn the hierarchical representation of graphs~\cite{Micha2016fastgcn,li2019deepgcns}, since GNNs (i) lack the effective and efficient pooling for coarsening unstructured graphs and (ii) also suffer from the gradient vanishing and over-smoothing problems when stacking more layers. As a result, this limitation makes GNNs especially problematic for graph classification (which associates a label with an entire graph). To address this problem, a hierarchical residual structure is adopted for our SGCN,  which  is capable of fully incorporating the local neighbourhood information and exploiting the global shape properties. To this end, the following two key ideas are introduced for our SGCN:
\begin{itemize}[leftmargin=*,itemsep=0.5pt]
\item Cluster-based Pooling~\cite{Micha2016fastgcn} derives a clustering on all the graph nodes and then aggregates the nodes of the same cluster with new computed coordinates, which eventually results in a coarsen graph.
\item Residual Learning~\cite{he2016deep,li2019deepgcns} adds shortcut connections between the input and output of graph convolutions, which empirically enables the reliably converge in training deep networks.
\end{itemize}
As shown in Fig.~\ref{fig:gcn_arch}~(a), the proposed SGCN follows the standard hierarchical residual structure  of conventional deep CNNs, where the figure~(b) details the spatial transform network (STN) and  (c) details the residual graph convolutional block (Rs-GCB). Additionally, each convolutional layer follows with the batch normalization~\cite{ioffe2015batch} and PReLU~\cite{he2015delving} for better convergence, and the dropout~\cite{Srivastava2014Dropout} is also utilized for good generalization. Finally, the SGCN is trained end-to-end by minimizing the $\mathbb{L}_2$ normalized cross-entropy~\cite{wang2018cosface}.

\subsection{Complexity Analysis of Graph Convolution}
Here we compare the computation complexity of convolutions on images and graphs for the task of OLHCCR. Generally, both images and graphs can be represented as graphs with their nodes and edges, i.e., $\mathbb{G}=(\mathbf{V}, \mathbf{E})$. Then, the complexity of a single channel convolution on a graph $\mathbb{G}$ should be $\mathcal{O}(|\mathbf{V}||\bar{\mathbf{E}}|)$, where $|\bar{\mathbf{E}}|$ denotes the average number of edges for each node, i.e., $|\bar{\mathbf{E}}|=\frac{1}{|\mathbf{V}|}\sum_{v_i \in \mathbf{V}}|\mathcal{N}(v_i)|$. Specifically, for an image with the height $H$ and weight $W$, we get $|\mathbf{V}|=HW$ and $|\bar{\mathbf{E}}|=9$ (i.e. eight neighbourhoods and itself); for a sparse character graph,  we can safely assume that  $|\bar{\mathbf{E}}|\approx3$ (i.e. the former point and subsequent points in temporal order and itself), since most of the nodes are not the intersection points. As a result, the convolution complexity comparison between the image and graph should be $r \approx \mathcal{O}(\frac{3NH}{|\mathbf{V}_g|})$, where $\mathbf{V}_g$ denotes the node-set of the graph. Empirically, to achieve comparable accuracy for OLHCCR, the resolution of an input image is typically fixed to $64\times64$, while a character graph only  contains nearly 100 nodes on average. On this condition, the convolution complexity of images is much more expensive than that of character graphs.

\section{Experiments}
\subsection{Datasets}
Online handwritten Chinese character datasets are as follows:
\begin{itemize}[leftmargin=*,itemsep=0.5pt]

\item{\textbf{IAHCC-UCAS2016}}~\cite{qu2018data} is a public in-air handwritten Chinese character dataset, where each character is written in the midair within a single stroke. The dataset contains totally 431,825 samples covering 3755 Chinese characters (level-1 set of GB2312-80), where each class contains 115 different samples. Similar to previous works~\cite{qu2018air}, 92 samples per class are chosen as the training set, and the remaining as the test set.

\item{\textbf{ICDAR-2013}}~\cite{yin2013icdar} is the most popular dataset of online handwritten Chinese characters that collected by the CASIA institution.  For the OLHCCR task, the sub-datasets CASIA-OLHWDB 1.0 \& 1.1  are used as the training set, which contains 2,693,183 samples; and the ICDAR-2013 competition sub-dataset is used as the test set, which contains 224,590 samples of 3755 character classes (level-1 set of GB2312-80).

\end{itemize}

\subsection{Implementation Details}
The whole architecture is entirely based on the PyTorch \cite{paszkepytorch} deep learning platform. In experiments, the detailed configuration of our spatial graph convolutional network for OLHCCR is shown in Fig.~\ref{fig:gcn_arch}, where the kernel size of each spline convolution is set as 3. Moreover, the dropout probability is set as 0.2 for each ``Rs-GCB'' module during training. The whole network is optimized via ADAM algorithm~\cite{kingma2014adam} with a batch size of 128. Furthermore, the initial learning rate is set at 0.002 and then decayed by $\times$ 0.1 when the performance stops improving. Finally, the training process is terminated when the model reaches the convergence. All the experiments are conducted on a Dell workstation with an Intel(R) Xeon(R) CPU E5-2630 v4 @ 2.20GHz, 32 GB RAM, and two  NVIDIA Quadro P5000 GPUs.

\subsection{Results with Varying Depths \& Widths}
\begin{table}[t]
\centering
\begin{tabular}{c|cc|ccc}
\toprule
\multirow{2}{*}{ID} &\multirow{2}{*}{Depth}&\multirow{2}{*}{Width}&Training&Storage& Accuracy\\
&&& (h)&(MB)&(\%)\\
\hline
\midrule
\#1&1&1.0$\times$&3.0&3.73&96.33\\
\#2&1&1.5$\times$&4.4&5.09&96.68\\
\#3&2&1.0$\times$&4.7&4.99 &96.84\\
\#4&2&1.5$\times$&7.6&7.92&96.99\\
\#5&3&1.0$\times$&6.6&6.17&96.85\\
\#6&3&1.5$\times$&10.4&10.56&96.98\\
\bottomrule
\end{tabular}
\vspace{-0.2cm}
\caption{Accuracies of SGCNs with varying block depths and widths on IAHCC-UCAS2016, i.g.``\#3'' corresponds to the SGCN in Fig.~\ref{fig:gcn_arch}.} \label{tab:vary_configs}
\vspace{-0.3cm}
\end{table}

To investigate the effectiveness of the proposed method for OLHCCR, we fully compare the performance of SGCNs with varying depths and widths as shown in Table~\ref{tab:vary_configs}. Additionally, we mainly search different network configurations on the small dataset (i.g. IAHCC-UCAS2016), hoping that the well-performed SGN architecture can be successfully transferred to the large dataset (i.g. ICDAR-2013). Specifically, during the comparison, different SGCNs follow the same structure as shown in Fig.~\ref{fig:gcn_arch} but with different numbers of convolutional layers and channels in each ``Rs-GCB'' module. For example, ``SGCN-\#3'' in Table~\ref{tab:vary_configs} is corresponding to the SGCN in Fig.~\ref{fig:gcn_arch}, and ``SGCN-\#4'' expands the width of ``SGCN-\#3'' by $\times$1.5. As shown in Table~\ref{tab:vary_configs}, either increasing the network depth or expanding its width can improve the recognition accuracy, however, both strategies correspondingly demand much more parameters and training time. Moreover, this accuracy improvement tends to become marginal when the model is sufficiently deep. In general, we prefer to choose the ``SGCN-\#3'' and ``SGCN-\#4'' for OLHCTR as the trade-off among the accuracy, storage, and training time.

\subsection{Ablation Study}
To fully analyze the effectiveness of each part in SGCN, we conduct the detailed ablation study of ``SCGN-\#3'' on IAHCC-UCAS2016 dataset as shown in Table~\ref{tab:ablation_study}. 

\begin{table}[t]
\centering
\vspace{-0.3cm}
\begin{tabular}{ccccc}
\toprule
\multicolumn{2}{l}{Feature}& \multicolumn{2}{l}{STN}&\multirow{2}{*}{Accuracy}\\
\cmidrule(l){1-2}\cmidrule(l){3-4}
Spatial&Temporal&Input&Feature&(\%)\\
\midrule
&&&&95.95$^{\downarrow\textbf{0.89}}$\\
\midrule
&&$\surd$&$\surd$&96.22$^{\downarrow\textbf{0.62}}$\\
$\surd$&&$\surd$&$\surd$&96.53$^{\downarrow\textbf{0.31}}$\\
&$\surd$&$\surd$&$\surd$&96.49$^{\downarrow\textbf{0.35}}$\\
\midrule
$\surd$&$\surd$&&&96.35$^{\downarrow\textbf{0.49}}$\\
$\surd$&$\surd$&$\surd$&&96.42$^{\downarrow\textbf{0.42}}$\\
$\surd$&$\surd$&&$\surd$&96.75$^{\downarrow\textbf{0.09}}$\\
\midrule
$\surd$&$\surd$&$\surd$&$\surd$&~\textbf{96.84}\\
\bottomrule
\end{tabular}
\vspace{-0.1cm}
\caption{Ablation study of  SGCN-$\#$3 on IAHCC-UCAS2016.} \label{tab:ablation_study}
\vspace{-0.15cm}
\end{table}

\begin{table}[t]
\centering
\begin{tabular}{lccc}
\toprule
\multirow{2}{*}{Method}&\multicolumn{3}{c}{Accuracy (\%)}\\
\cmidrule(lr){2-4}
&{1a}&{1b}&{1c}\\
     \midrule
    HMM + DTW&~\multirow{2}{*}{97.1}& ~\multirow{2}{*}{92.8}& ~\multirow{2}{*}{90.7}\\
        \addlinespace[-0.65ex]
    ~~\cite{Bahlmann2004thew}&&&\\
    LSTM& ~\multirow{2}{*}{96.84}& ~\multirow{2}{*}{92.31}& ~\multirow{2}{*}{89.79}\\
    \addlinespace[-0.65ex]
    ~~\cite{Hochreiter1997Long}&&&\\
    1D-CNN& ~\multirow{2}{*}{98.08}& ~\multirow{2}{*}{94.67}& ~\multirow{2}{*}{{95.33}}\\
    \addlinespace[-0.65ex]
    ~~\cite{Iwana2019dwa}&&&\\
        DWA 1D-CNN& ~\multirow{2}{*}{98.54}& ~\multirow{2}{*}{96.08}& ~\multirow{2}{*}{\textbf{95.92}}\\
    \addlinespace[-0.65ex]
    ~~\cite{Iwana2019dwa}&&&\\
    Google Hybrids&~\multirow{2}{*}{\textbf{99.2}}& ~\multirow{2}{*}{\textbf{96.9}}& ~\multirow{2}{*}{94.9}\\
    \addlinespace[-0.65ex]
    ~~\cite{Keysers2017multi}&&&\\
\midrule
     SGCN [Ours] &\emph{98.76} &\emph{96.33}&\emph{95.52} \\
    \bottomrule
\end{tabular}

\vspace{-0.15cm}
\caption{Classification accuracies of digits and letters on UNIPEN.} \label{tab:unipen_results}
\vspace{-0.35cm}
\end{table}

\paragraph{Node Feature Analysis}
We first analyze the effectiveness of different node features. As shown in Table~\ref{tab:ablation_study}, if no features are provided for graph nodes (i.e. features of all nodes are set into the same value), the SGCN still achieves a slightly good accuracy. This indicates that the graph representations well retain the geometric properties of characters, and the spatial graph convolution can also exploit their spatial structures via the local neighbourhoods aggregation and hierarchical structure learning. Moreover, the recognition accuracy will increase if we explicitly assign either spatial features (i.e. the absolute coordinates) or temporal features (i.e. writing offsets \& directions) to the graph nodes. Finally, the SGCN achieves the best result by combining both spatial and temporal features, demonstrating the necessity of feature extraction.

\paragraph{Spatial Transformation Analysis} We further analyze the effectiveness of spatial transformation~(ST) for inputs and features.  We notice that ST for inputs does not bring a noticeable accuracy improvement for character recognition, which reveals that (1) variations in scales and positions can be effectively addressed by pre-processing and (2) the SGCN is robust to the small rotations of characters. On the contrary, a significant accuracy increase is observed when the ST is adopted in the feature space. This may indicate that the ST transforms high dimensional features into the latent canonical forms, and thus it helps learn the geometric invariant features. This eventually benefits the final classification.

\paragraph{Extension to Digits and Letters} To demonstrate that the SGCN is language-independent, we evaluate the SGCN on the UNIPEN dataset ~\cite{guyon1994unipen}, which consists of the isolated digits (\textbf{1a}), upper case (\textbf{1b}) and lower case English letters (\textbf{1c}) respectively. Moreover, the detailed configuration of SGCN for UNIPEN is ``STN $\rightarrow$ FeatLayer $\rightarrow$ Rs-GCB(32) $\rightarrow$ Pooling $\rightarrow$ STN $\rightarrow$ Rs-GCB(64) $\rightarrow$ Pooling  $\rightarrow$ Global-Average$\rightarrow$FC(128) $\rightarrow$ Softmax''. As shown in Table~\ref{tab:unipen_results}, the SGCN achieves comparable accuracies compared with previous methods on the UNIPEN datasets. This indicates that the SGCN is not limited to Chinese characters and it also can be extended to other languages easily.


\subsection{Benchmarking Results}
\vspace{-0.1cm}
\begin{table}[t]
\centering
\vspace{-0.3cm}
\begin{tabular}{rlrr}
\toprule
\multirow{2}{*}{Method}& \multirow{2}{*}{Ref.}&Storage&Accuracy\\
&&(MB)& (\%)\\
\midrule
    MQDF&\cite{liu2013online}&188.35 & 89.96\\
    LSROPC&\cite{qu2018air}& 68.04& 91.01\\
\addlinespace[0.65ex]

    DMap-CNN&\cite{qu2018data}& 20.2& 92.81\\
    2D-CNN&\cite{gan2020comp}& 15.5& 95.33\\
    \addlinespace[0.65ex]
        GRU&\cite{ren2017end}& \emph{7.03} &92.50 \\
    Attn-GRU&\cite{ren2019recog}&8.71 &93.18 \\
    1D-CNN&\cite{gan2018anew}& 8.09& {96.78}\\
\midrule
    SGCN-\#3 &[Ours] &{\textbf{4.99}} &\emph{96.84}\\
    SGCN-\#4 &[Ours] &{7.92} &{\textbf{96.99}}\\
\bottomrule
\end{tabular}
\vspace{-0.2cm}
\caption{Comparison of classification accuracies for in-air handwritten Chinese characters on IAHCC-UCAS2016.} \label{tab:iahcc_results}
\vspace{-0.2cm}
\end{table}

\begin{table}[t]
\centering
\begin{tabular}{rlrr}
\toprule
\multirow{2}{*}{Method}& \multirow{2}{*}{Ref.}& Storage&Accuracy\\
&&(MB)& (\%)\\
\midrule
Human&\cite{yin2013icdar}& N/A & 95.19\\
DLQDF&\cite{liu2013online}& 120.0 & 95.31 \\
\addlinespace[0.65ex]
VO3-CNN &\cite{yin2013icdar}& 87.60 & 96.87\\

DS-CNN&\cite{yang2016dropsample}&15.0  & 97.23\\
UWarwick &\cite{yin2013icdar}& 37.80 & 97.39\\
DMap-CNN& \cite{zhang2017online}& 23.50 &97.55  \\
 \addlinespace[0.65ex]
 Attn-GRU&\cite{ren2019recog}&\emph{7.01}&97.30\\
LSTM&\cite{zhang2018drawing} &10.38 & \emph{97.76}\\
1D-CNN&\cite{gan2018anew} &8.05& \textbf{97.86}\\
\midrule
SGCN-\#3 &[Ours] &\textbf{4.99} &{97.34}\\
SGCN-\#4 &[Ours] &{7.92} &{97.55}\\
\bottomrule
\end{tabular}
\vspace{-0.2cm}
\caption{Comparison of classification accuracies for online handwritten Chinese characters on ICDAR-2013.} \label{tab:icdar_results}
\vspace{-0.3cm}
\end{table}

Lastly, we fully compare the SGCN with previous methods on benchmark datasets (including IAHCC-UCAS2016 and ICDAR-2013) as listed in Table~\ref{tab:iahcc_results} and Table~\ref{tab:icdar_results}. Particularly, we notice that the SGCN on graphs requires much fewer parameters than the 2D-CNN on feature images to achieve comparable accuracy. Moreover, the SGCN also achieves comparable accuracies and similar storages compared with the 1D-CNN or RNN on temporal features; however, the accuracy of the SGCN does not heavily rely on the temporal information (as shown in Table~\ref{tab:ablation_study}), while the 1D-CNN and RNN will not work for classification when temporal orders of trajectories are lacked or disturbed. Overall, experiments demonstrate that the SGCN is comparable with the state-of-the-art methods, yet the SGCN not only has its unique advantages but also addresses OLHCCR in a completely different perspective . 

\vspace{-0.25cm}
\section{Conclusion}
\vspace{-0.15cm}
In this paper, we have proposed a novel spatial graph convolutional network (SGCN) for OLHCCR, upon which characters are viewed as geometric graphs. This is largely different from the latest popular methods including the 2D-CNN, 1D-CNN, and LSTM, thus addressing character recognition in a novel perspective. Experiments on benchmarks demonstrate the effectiveness of SGCN for OLHCCR. In future work, we plan to (1) design new graph convolution kernels for better performance and (2) extend our SGCN to offline HCCR. 

%

\small
\bibliographystyle{named}
\bibliography{arxiv}

\end{document}